\documentclass[lettersize,journal, onecolumn,draftclsnofoot]{IEEEtran}
\usepackage{lineno}

\usepackage{amsmath,amsfonts}
\usepackage[english]{babel}
\usepackage{hyperref}
\hypersetup{
    colorlinks=true,
    linkcolor=blue,
    citecolor=blue,
    filecolor=magenta,      
    urlcolor=cyan,
    pdfpagemode=FullScreen,
    }

\usepackage{algorithmic}
\usepackage{algorithm}
\usepackage{afterpage}
\usepackage{array}
\usepackage[caption=false,font=normalsize,labelfont=sf,textfont=sf]{subfig}
\usepackage{xcolor}
\usepackage{longtable}
\usepackage{booktabs} 
\usepackage{textcomp}
\usepackage{stfloats}
\usepackage{url}
\usepackage{verbatim}
\usepackage{graphicx}
\usepackage{csquotes}
\usepackage{subcaption}
\usepackage{graphicx} 
\usepackage{tikz}
\usepackage{multirow}
\usepackage{pdflscape}
\usepackage{cite}

\newcolumntype{C}[1]{>{\centering\let\newline\\\arraybackslash\hspace{0pt}}m{#1}}

\usepackage{siunitx}

\newcolumntype{C}[1]{>{\centering\let\newline\\\arraybackslash\hspace{0pt}}m{#1}}
\usepackage{booktabs} 

\addto\extrasenglish{%
}

\begin{document}

\title{Deep Learning for Human Locomotion Analysis in Lower-Limb Exoskeletons: A Comparative Study}

\author{Omar Coser$^{1,2}$, Christian Tamantini$^{2,5}$, Matteo Tortora$^{1,3}$, Leonardo Furia$^{1}$, Rosa Sicilia$^{1}$, Loredana Zollo$^{2}$, Paolo Soda$^{1,4}$
\thanks{$^{*}$ Corresponding author: O. Coser (email:  omar.coser@unicampus.it)}\\
\thanks{$^{1}$Unit of Computer Systems \& Bioinformatics, Università Campus Bio-Medico di Roma, Via Álvaro del Portillo, 21, Rome, 00128, Italy. $^{2}$Unit of Advanced Robotics and Human-Centered Technologies, Università Campus Bio-Medico di Roma, Via Álvaro del Portillo, 21, Rome, 00128, Italy. 
$^{3}$Dipartimento di Ingegneria Navale, Elettrica, Elettronica e delle Telecomunicazioni, Università degli Studi di Genova, Via all’Opera Pia 11a, 16145 Genova, Italy.
$^{4}$Department of Diagnostics and Intervention, Radiation Physics, Biomedical Engineering, Umeå University, Universitetstorget, 4, Umeå, 40196, Sweden.
$^{5}$Institute of Cognitive Sciences and Technologies, National Research Council of Italy, 00196 Rome, Italy}
\thanks{This work was supported by the Italian Ministry of Research, under the complementary actions to the NRRP  ``Fit4MedRob - Fit for Medical Robotics'' Grant (PNC0000007, CUP: B53C22006990001). Authors thank prof. G. Iannello for his support in the research. Submitted on Frontiers in Computer Science.} }



\maketitle

\begin{abstract}
Wearable robotics for lower-limb assistance have become a pivotal area of research, aiming to enhance mobility for individuals with physical impairments or augment the performance of able-bodied users. Accurate and adaptive control systems are essential to ensure seamless interaction between the wearer and the robotic device, particularly when navigating diverse and dynamic terrains. 
Despite the recent advances in  neural networks for time series analysis, no attempts have been directed towards the classification of ground conditions, categorized into five classes and subsequently determining the ramp's slope and stair's height.
In this respect, this paper presents an experimental comparison between eight deep neural network backbones to predict high-level locomotion parameters across diverse terrains. 
 All the models are trained on the publicly available CAMARGO 2021 dataset.
IMU-only data equally or outperformed IMU+EMG inputs, promoting a cost-effective and efficient design. Indeeds, using three IMU sensors, the LSTM achieved high terrain classification accuracy (0.94 ± 0.04) and precise ramp slope (1.95 ± 0.58°) and the CNN-LSTM a stair height (15.65 ± 7.40 mm) estimations. As a further contribution, SHAP analysis justified sensor reduction without performance loss, ensuring a lightweight setup. The system operates with ~2 ms inference time, supporting real-time applications. The code is code available at https://github.com/cosbidev/Human-Locomotion-Identification.

\end{abstract}

\begin{IEEEkeywords}
Deep Learning; Explainable AI; Human Locomotion; Multimodal Learning; Neural Networks
\end{IEEEkeywords}

\section{Introduction}
\label{soa2}
The field of lower limb robotics for rehabilitation and assistance leverages advanced robotic technologies to support individuals with lower limb impairments or disabilities~\cite{diaz2011lower}. Across clinical practice, robotic applications range from wearable exoskeletons designed to augment natural movement~\cite{lee2020lower} and robotic prostheses that restore ambulation in users with artificial limbs~\cite{o2019prosthetic}, to rehabilitation robots that deliver targeted exercises~\cite{valgeirsdottir2022we}. By providing enhanced rehabilitation, fostering independence, and enabling personalized therapeutic solutions, these innovations hold significant promise for improving patient outcomes. However, optimal performance in these systems requires robust data-driven parameter extraction and context-aware interventions. Specifically, control strategies in lower limb robotic systems must incorporate gait parameters to dynamically adapt to varying terrain conditions~\cite{gao2023autonomous}.

Human locomotion analysis—encompassing terrain recognition and quantification of slope or stair height—plays a pivotal role in lower limb robotics, particularly in assistive devices such as prostheses and exoskeletons. Wearable sensors have become indispensable for capturing locomotion data and providing detailed insights into human movement. Two of the most frequently used sensor types are Inertial Measurement Units (IMUs), which measure acceleration and angular velocity to capture body orientation and movement~\cite{ribeiro2017inertial, bartlett2017phase}, and electromyography (EMG) sensors, which detect muscle activity to illuminate neuromuscular function and movement patterns~\cite{gupta2017semg, sorkhabadi2019human}.

Traditional machine learning techniques—such as k-Nearest Neighbor (kNN), Support Vector Machines (SVM), Gaussian Mixture Models (GMM), and Random Forests (RF)—have been widely applied for time-series classification of locomotion data from IMU and EMG sensors. Despite their effectiveness, these methods rely on manual feature engineering, a process that is labor-intensive and requires significant domain expertise~\cite{attal2015physical}. Consequently, a growing body of research has shifted toward deep learning approaches that leverage neural networks (NNs) for automated feature extraction and classification.

Table~\ref{tab:soa} offers a comparative overview of existing methodologies in human locomotion analysis, focusing on sensor modalities, algorithms, datasets, classification tasks, and validation approaches. It encompasses a range of studies that use EMG (electromyography), IMU (inertial measurement units), or both as data sources, and it showcases both traditional machine learning models like LDA, kNN, DT, RF, and SVM~\cite{negi2020human, zhang2020unsupervised, zheng2022cnn} and advanced deep learning architectures such as CNN, LSTM, hybrid CNN-LSTM models, ResNet, Artificial Neural Network (AAN), Domain-Adversarial Neural Network(DANN), Maximum Classifier DIscrepancy (MCD) and Attention-based CNN-BiLSTM~\cite{narayan2021real, amer2021human, wang2022locomotion, zhao2022multi, jing2022accurate, liang2023deep, kang2022subject, le2022deep}. Each study briefly outlines its objectives and methodology, addressing tasks such as gait analysis, locomotion intent prediction, real-time mode recognition, joint moment estimation, and terrain classification. Reported accuracies generally exceed 90\%, including 0.99 accuracy using CNN with IMU data and 0.94–0.95 accuracy with CNN-LSTM models. Lower accuracy outcomes (around 0.74) are primarily observed in transfer learning scenarios. Public datasets like CAMARGO~\cite{camargo2021comprehensive}, ENABL3S~\cite{hu2018benchmark}, and DSADS~\cite{barshan2014recognizing}, along with custom datasets, cover between four and ten classes (e.g., level ground, ramp ascent/descent, stair ascent/descent, and additional activities such as sitting and standing), with participant counts varying from five to 500 subjects. Validation strategies include k-fold cross-validation, leave-one-subject-out, hold-out, and leave-one-terrain-out, influencing model generalizability. IMU-based models typically exhibit strong real-time performance, whereas multimodal setups offer varied benefits. Hybrid architectures like CNN-LSTM tend to outperform simpler models, particularly for complex movement tasks. Dataset choice and validation methods critically affect model outcomes.

Despite these notable advancements, several challenges persist. Much of the existing literature concentrates on unimodal analysis, with fewer studies providing a comprehensive comparison of architectures and sensor combinations or investigating sensor reduction strategies to maintain performance while minimizing complexity. Table~\ref{tab:soa} underscores this gap by noting a clear predominance of unimodal approaches, with only a handful of studies considering multimodal analyses~\cite{negi2020human, zhang2020unsupervised, zhao2022multi}.

Furthermore, the majority of neural-network-based solutions for lower limb robotics concentrate on classification tasks, such as locomotion mode detection, and often neglect predictive aspects like ramp inclination or step height estimation. Moreover, only a few studies address the minimal number of sensors necessary for robust performance a critical consideration for real-world deployments where system cost and complexity must be balanced with performance requirements.

In this paper, we address these research gaps by proposing a neural-network-based system that not only classifies ground conditions (level ground, stair ascent, stair descent, ramp ascent, and ramp descent) but also predicts ramp inclination or step height using multimodal signals from the CAMARGO dataset. Our approach aims to enhance adaptability, reduce sensor dependency, and improve real-world applicability in lower limb robotics. Specifically: 
\begin{enumerate} 
\item We present a robust comparative analysis of eight models commonly used for time series classification and regression within the domain of multimodal human locomotion. Deep learning techniques were chosen due to their ability to automatically learn complex patterns and features directly from raw data, eliminating the need for manual feature extraction, which is often time-consuming, domain-dependent, and prone to human bias. Deep learning models, particularly in time series tasks, excel at capturing temporal dependencies and non-linear relationships, leading to improved predictive performance A leave-one-subject-out cross-validation approach is employed to ensure generalizability. 
\item We compare multimodal and unimodal setups to determine whether a multimodal configuration offers any advantage over a unimodal approach. Using Explainable AI techniques, we identify the most influential features—here, corresponding to unimodal sensor signals—and then conduct an ablation study to establish the minimal sensor configuration that achieves optimal performance. 
\item We used the CAMARGO 2021 dataset that offers a comprehensive, high-quality, and ethically sound foundation for human locomotion recognition research.
\end{enumerate}
The rest of the manuscript is organized as follows. Section~\ref{sc:materials} describes the dataset and the adopted pre-processing methods. Section~\ref{sc:methods} elaborates on the proposed neural network approaches and the experimental setup. Section~\ref{sc:results} presents and discusses the results, while Section~\ref{sc:concl} provides concluding remarks and outlines potential future research directions.
\begin{landscape}
\begin{longtable}{c C{1.5cm} C{1.5cm} C{8cm} C{1cm} C{2cm} C{2cm} C{2cm} C{2cm}}
\caption{Summary of methodologies for human locomotion analysis for terrain and slope recognition. (LG:Level Ground, RA/D:Ramp Ascent/Descent, SA/D: Stair Ascent/Descent), * indicate multimodal analysis}\label{tab:soa}\\
\toprule
\textbf{Ref.} & \textbf{Modality} & \textbf{Algorithm} & \textbf{Brief description} & \textbf{Acc} & \textbf{Dataset} & \textbf{\# of classes} & \textbf{\# of subject} & \textbf{Validation approach}\\
\midrule
\endfirsthead

\multicolumn{9}{c}%
{{\bfseries Table \thetable\ (continued)}} \\
\toprule
\textbf{Ref.} & \textbf{Modality} & \textbf{Algorithm} & \textbf{Brief description} & \textbf{Accuracy} & \textbf{Dataset} & \textbf{\# of classes} & \textbf{\# of subject} & \textbf{Validation approach}\\
\midrule
\endhead

\midrule
\multicolumn{9}{r}{{Continued on next page}} \\
\endfoot

\bottomrule
\endlastfoot
\cite{negi2020human} &  EMG and acceleration, * early fusion       &  kNN, SVM, RF, LDA, DT  &Gait analysis using surface electromyography and acceleration sensors classified five terrains. Signals from tibialis anterior and gastrocnemius muscles were processed with machine learning models. The goal was to optimize classification accuracy with minimal computation time and muscle signals.& 0.97, 0.98, 0.79, 0.80, 0.99    &  Collected within the paper (CWP)  &5 classes (LG, RA, RD, SA, SD) &15 subjects & k-fold-cross-validation   \\\hline
\cite{zhang2020unsupervised} &    EMG and IMU, * early fusion      &     LDA, SVM, ANN, CNN, DANN, MCD   & Predicting human locomotion intent aids in controlling wearable robots and assisting movement on various terrains. This study introduces an unsupervised cross-subject adaptation method to predict locomotion intent without labeled data.  &    0.92, 0.90, 0.93, 0.96, 0.95, 0.95     & ENABL3S \cite{hu2018benchmark} and DSADS \cite{barshan2014recognizing} & 5 classes LG, SA, SD, RA, RD & 10 subjects &leave-one-subjects out \\\hline
\cite{narayan2021real}  &    IMU      &     CNN   & This study evaluates hierarchical classification for real-time locomotion mode recognition in wearable robotic prostheses and exoskeletons. A CNN-based classifier trained on inertial sensor data achieves stable and accurate mode classification, including smoother transitions. The method enables real-time operation, improving control adaptation in wearable robots.  &     0.94     &   CWP & 10 classes (sit, stand, walk str, walk cur-left, walk cur-right, down s, up s, left t, right t, unknown)& 8 subjects& k-fold cross validation  \\\hline
\cite{amer2021human} &    IMU      &     CNN   & This study introduces a CNN-based algorithm for classifying human locomotion activities using inertial measurement unit (IMU) data. Spectral analysis transforms inertial signals into time-frequency representations, which are then classified as images.   &     0.99     &    CWP& 6classes (Norm walking, Walking upstairs, Walking downstairs, Sitting, Standing, Laying& 30 subjects&k-fold cross validation    \\\hline
\cite{wang2022locomotion} &    IMU      &     CNN  &  A residual network-based method is proposed for locomotion mode recognition in lower limb exoskeletons. Using inertial sensor data, the network autonomously learns mixed features, eliminating the need for manual feature extraction.  &     0.97    &   CWP  &5 classes LG, SA, SD, RA, RD&5 subjects& leave one subjects out  \\\hline
\cite{zhao2022multi} &    EMG and IMU,* early fusion       &    CNN  &A multi-channel separated encoder-based convolutional neural network is proposed for locomotion intention recognition. Inertial sensor data are processed through spectral transformation and classified using CNNs to enhance recognition accuracy.  &     0.94     &   ENABL3S \cite{hu2018benchmark} &5 classes, LG, RA, RD, SA, SD&10 subjects& k-fold cross validation  \\\hline
\cite{jing2022accurate} &    EMG      &    CNN  &  A neural network trained on HDsEMG data achieved higher accuracy and robustness against electrode shifts compared to bipolar electrodes. The approach enhances gait mapping reliability, supporting real-time control of assistive technologies.   &     0.97  &   HDsEMG \cite{jiang2021open} &6 classes, Standing, LG, SA, SD, RA, RD&7 subjects&k-fold cross validation\\\hline
\cite{zheng2022cnn} &    IMU      &     CNN-SVM  & A hybrid CNN–SVM model is proposed for locomotion mode recognition using multi-channel inertial measurement unit (IMU) signals. The approach integrates a feature mapping layer with error correction from a finite state machine (FSM) to improve accuracy and generalization.   &     0.98     &   CWP &LW, SA, SD, RA, RD&10 subjects&Leave one subject out \\\hline
\cite{le2022deep} &    IMU      &     CNN  &A deep convolutional neural network is developed for locomotion intent prediction in powered prosthetic legs, with both subject-dependent and subject-independent validations. Transfer learning is applied to improve subject-independent performance using a small portion of data from a new subject, significantly reducing error rates.   &     0.74     &   ENABL3S\cite{hu2018benchmark} &5 classes, LG, SA, SD, RA, RD&9 subjects&hold-out    \\\hline
\cite{kang2022subject} &    IMU      &   CNN   & A deep convolutional neural network-based locomotion mode classifier is developed for hip exoskeletons using an open-source gait biomechanics dataset. The model operates independently of user-specific data, ensures smooth mode transitions, and relies only on minimal wearable sensors. &     0.93     &  CAMARGO \cite{camargo2021comprehensive} &5-classes LG, SA, SD, RA, RD&21 subejcts&leave-one-terrain-out    \\\hline
\cite{son2023multivariate} &    EMG      &  CNN-LSTM, LSTM-CNN  & Data from electromyograms (EMGs) and robot sensors were used to compare the performance of two hybrid model LSTM-CNN and CNN-LSTM  &     0.94, 0.95     &   CWP&5-classes LG, SA, SD, RA, RD&500 subjecs&hold-out    \\\hline
\cite{liang2023deep} &   IMU      &  Attention-based CNN-BiLSTM &A deep-learning approach is developed for estimating lower-limb joint moments during locomotive activities using inertial measurement units (IMUs). The model accurately predicts hip, knee, and ankle joint moments with a single IMU, with the shank identified as the optimal placement. & 0.85 & CAMARGO \cite{camargo2021comprehensive} &4 classes: LG, Ramp, Stair, Treadmill&19 subjects& hold-out \\

\end{longtable}
\end{landscape}

\section{Materials}
\label{sc:materials}
\begin{figure}[H]
    \centering
    \includegraphics[width=10cm]{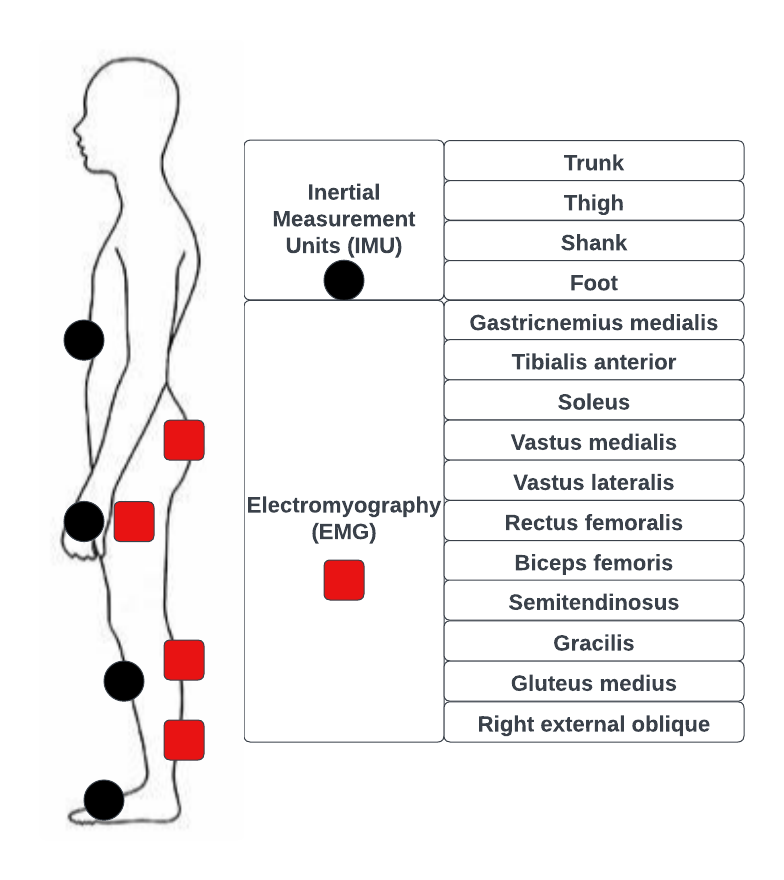}
    \caption{Position of the sensors along the body (Image extracted \cite{camargo2021comprehensive})}
    \label{fig:sensors}
\end{figure}

In this work, we utilize the CAMARGO dataset~\cite{camargo2021comprehensive}, a publicly available multimodal repository including sensor data from 21 subjects. 
Each participant was equipped with four IMUs positioned on the trunk, thigh, shank, and foot, along with 11 EMG sensors that monitored the activities of the gastrocnemius medialis, tibialis anterior, soleus, vastus medialis, vastus lateralis, rectus femoris, biceps femoris, semitendinosus, gracilis, gluteus medius, and right external oblique muscles.
The participants completed multiple trials across five locomotion modes: level ground walking, ramp ascent and descent, and stair ascent and descent. 
Stair trials were performed at four different heights (\SI{102}{\milli\meter}, \SI{127}{\milli\meter}, \SI{152}{\milli\meter}, \SI{178}{\milli\meter}), while ramp trials included six inclination angles (\ang{5.2}, \ang{7.8}, \ang{9.2}, \ang{11}, \ang{12.4}, and \ang{18}). 
In particular, the sensors are positioned along the body as shown in Figure \ref{fig:sensors}, We can see that there are eleven Electromyography (EMG) sensors on the right side, targetting major lower limb muscles and four Inertial Measurement Units (IMUs) attached to the torso, thigh, shank and foot, capturing acceleration and angular velocity, For a total of 35 signals recorded over an estimated 40 minutes, capturing high-resolution IMU and EMG data across various locomotion conditions. For a more detailed description of a dataset and the exact sensor placements, please refer to the original paper \cite{camargo2021comprehensive}

The authors of the dataset preprocessed the collected dataset: whereas the IMU data, comprising acceleration and angular velocity from the three-axis accelerometer and gyroscope sampled at \SI{200}{\hertz}, were processed using a lowpass filter with a \SI{100}{\hertz} cutoff frequency (Butterworth order 6), the raw EMG data, sampled at \SI{1000}{\hertz}, was digitally conditioned using a bandpass filter with a cutoff frequency of \SI{20}{\hertz} to \SI{400}{\hertz} (Butterworth order 20). 

In our work, we use the data as described so far, utilizing a rolling window approach to segment the data into non-overlapping temporal windows of \SI{500}{\milli\second} to generate approximatly 5000 samples for each subject with a class probability of $20 \pm 2.5\%$ (the number of samples and the class probability depend slightly on the subject) to feed the different NN models. 

\section{Methods}
\label{sc:methods}
\begin{figure}[H]
    \centering
    \includegraphics[width=18cm]{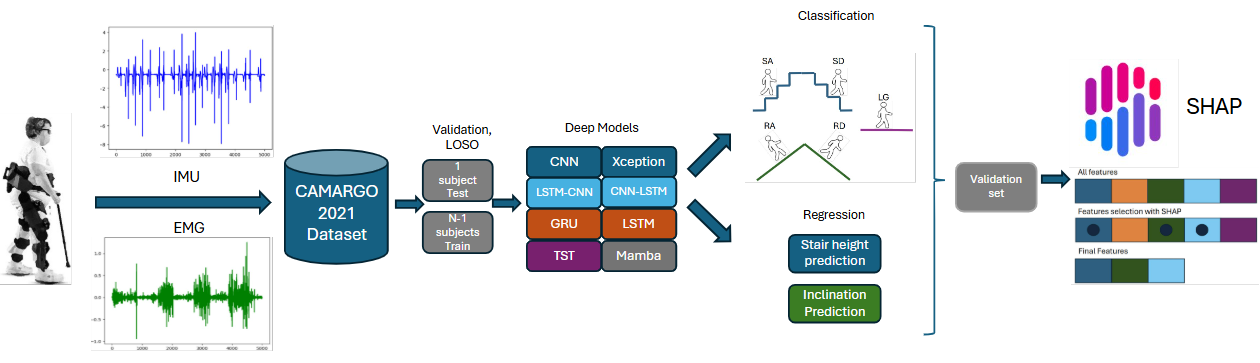}
    \caption{Visual Methodology}
    \label{fig:galaxy}
\end{figure}

This section outlines the methodology for developing a system architecture to classify terrain types and estimate terrain parameters using multimodal sensory data. 
The system comprises two separate stages: a classification network for detecting the five terrain types and two regression models to estimate the slope of a ramp or the height of a stair.
We compared state-of-the-art DL-based models for time series analysis to identify the optimal configuration. 
We selected the best-performing sensor modality and model architecture based on the results. 
Additionally, we used Explainable AI (XAI) techniques to determine the most influential features, providing insights into sensor importance and the minimal sensor setup required for accurate classification and regression as shown in Figure \ref{fig:galaxy}.

\subsection{Supervised Models for Human Locomotion Parameters Identification}
In this study, we evaluated eight core deep learning architectures,  in particular those models can be categorized in five major class, the Convolutional Neural Network (CNN, XceptionTime), hybrid models (CNN-LSTM and LSTM-CNN), Recurrent neural network (LSTM and Gated Recurrent Unit), Transformer (TST) and a variant of RNN the State Space Model (MAMBA). They are:
\begin{itemize}
    \item\textbf{Convolutional Neural Networks (CNN, XceptionTime)}: CNNs \cite{kang2022subject} and XceptionTime \cite{rahimian2019xceptiontime} are renowned for their ability to efficiently extract local and hierarchical features from time series data. While CNN provides a solid baseline for feature extraction, XceptionTime enhances this capability with depthwise separable convolutions and parallel temporal paths, improving computational efficiency and multi-scale pattern recognition.
    \item \textbf{Hybrid Models (CNN-LSTM, LSTM-CNN)}: Combining convolutional and recurrent layers allows hybrid models to leverage both spatial feature extraction and temporal sequence modeling. CNN-LSTM \cite{son2023multivariate} first captures local patterns using CNNs before modeling temporal dependencies with LSTMs, while LSTM-CNN \cite{son2023multivariate} reverses this flow to prioritize sequential information before feature extraction. These models are particularly effective in tasks where both spatial and temporal structures are crucial.
    \item \textbf{Recurrent Neural Netoworks (LSTM, GRU)}: LSTM (custom) and GRU \cite{chung2014empirical} are well-established architectures for handling sequential data. LSTMs excel at capturing long-term dependencies through their gating mechanisms, while GRUs offer a simplified structure with comparable performance and reduced computational overhead. Their inclusion ensures a solid benchmark for traditional sequence modeling approaches.
    \item  \textbf{Transformer-Based Model (TST)}: The Time Series Transformer (TST) \cite{zerveas2021transformer} introduces self-attention mechanisms to time series analysis, enabling the model to capture long-range dependencies without relying on recurrent structures. This approach enhances parallelization and scalability, addressing the limitations of RNN-based models in handling large and complex datasets.
    \item \textbf{State Space Model (MAMBA)}: MAMBA \cite{gu2023mamba} represents a modern variant of RNNs, leveraging structured State Space Models to efficiently capture long-range dependencies and continuous-time dynamics. Its inclusion allows for the exploration of state-of-the-art techniques in time series modeling, particularly for datasets with irregular intervals or complex temporal relationships.
\end{itemize}
These architectures encompass a wide range of modeling techniques, from convolutional approaches focused on local feature extraction to recurrent and state-space models designed for capturing temporal dependencies, and finally, transformer-based models that excel at modeling global relationships within sequences. Depending on the modality, the models take as input n features, leveraging multimodal strategies to enhance predictive performance. Among these strategies, early fusion is frequently employed, where raw sensor data is merged into a shared embedded space via concatenation, addition, pooling, or gated units~\cite{tortora2023radiopathomics}. Our choice to use early fusion follows from the data’s homogeneous nature and consistent sampling frequencies across modalities, eliminating the need for complex alignment or resampling. These uniform data characteristics allow the model to capture meaningful cross-modal interactions from the outset, enhancing predictive capability while minimizing noise or redundancy. Additionally, early fusion offers a simpler and more efficient architecture by removing the requirement for separate processing pathways for each modality, thereby reducing computational overhead. While multimodal sensor fusion can pose challenges—such as managing redundancies and handling missing modalities—the inherent synchronization and similarity in our data mitigate these issues, making early fusion an optimal choice for effective feature integration.

\subsection{Explainable AI}

Understanding and interpreting machine learning models is essential for ensuring transparency, trust, and fairness in AI-driven decision-making. There are various interpretability methods, categorized based on different aspects. Model-specific methods leverage the internal structure of models, whereas model-agnostic approaches, like SHAP, treat models as black boxes \cite{molnar2020interpretable, lundberg2017unified, ribeiro2016should}. Global explanations analyze overall model behavior, while local explanations focus on individual predictions \cite{doshi2017towards, ribeiro2016should}. Post-hoc methods provide insights without altering the model, whereas intrinsically interpretable models (e.g., decision trees) are inherently transparent \cite{guidotti2018survey, lipton2018mythos}. Feature attribution techniques, such as SHAP and LIME, quantify the impact of input features \cite{lundberg2017unified, ribeiro2016should}, while example-based explanations use representative instances to justify predictions \cite{caruana1999case, molnar2020interpretable}. Finally, visualization techniques, including Grad-CAM and saliency maps, enhance interpretability in deep learning by highlighting influential input regions \cite{simonyan2013deep, selvaraju2017grad}. These diverse methods play a crucial role in making AI systems more understandable and accountable.
SHAP was chosen because it is specifically well-suited for explaining deep learning models used in time series tasks.
It provides theoretically grounded feature attributions based on Shapley values, ensuring local accuracy, consistency, and reliability.
In LSTMs and CNN-LSTMs, each time step can be treated like a separate “player,” capturing how each segment influences the overall prediction.
Its model-agnostic framework simplifies integration with any Python-based deep learning library without requiring internal access to the models.
This attribute is critical when dealing with complex multi-layer architectures, as in the case of time series forecasting and classification.
Moreover, SHAP’s clear visualization tools help pinpoint which periods or features are most influential, a vital need in time series analysis.
Compared to other interpretability tools, SHAP enforces consistency, meaning that more influential features receive higher attributions.
Finally, SHAP’s Python implementation is straightforward, making it a practical and efficient choice for researchers and practitioners alike.

\subsection{Experimental Setup}
We validated all experiments using the Leave-One-Subject-Out (LOSO) cross-validation method, conducting a number of runs equal to the number of subjects in our study 21. 
In each run, the training set included data from \(N-1\) subjects, while the test set consisted of data from the remaining subject.
We limited each model’s training to 10 epochs in total and added an early stopping rule, which would end training if the validation loss stopped improving for 100 consecutive epochs. In practice, the models typically stopped progressing well before hitting 100 consecutive epochs without improvement, so training naturally concluded at the 10th epoch once it became evident that the validation loss was no longer decreasing. Essentially, by the time we reached 10 epochs, the network had plateaued in terms of performance, triggering the early stopping condition and preventing any further, unproductive epochs.
All experiments and model training were performed on Google Colab Pro,  utilizing 52 GB of RAM (CPU) and a 15 GB GPU, except for Mamba that was trained on an NVIDIA A100 GPU with 80Gb of Ram resources.
For the classification task, we assessed performance using accuracy, precision, recall, F1 score, given the balanced nature of the dataset. 
However, for regression tasks, we evaluated performance using the Mean Absolute Error (MAE). 
To further assess model performance under different conditions, we performed the Wilcoxon signed rank test, a nonparametric statistical hypothesis test~\cite{gehan1965generalized}, and applied the Bonferroni correction to address the issue of multiple comparisons~\cite{napierala2012bonferroni}.

\subsection{Performance Indicators}

In this section, we present the main results of the experiments conducted. The primary metric used for evaluating classification performance is accuracy, used because the a priori class distribution is balanced. However, additional metrics, including recall and the F1 score, are reported in the Supplementary Material.
Furthermore, for the regression analysis, we employed the Mean Absolute Error (MAE) as the primary metric.

Additionally, to determine the contribution of each sensor to the model’s predictions, we computed the SHAP (SHapley Additive exPlanations) values \cite{lines2012shapelet}, enabling an in-depth assessment of sensor importance and model interpretability. This analysis provided insights into the relative influence of each sensor on the overall system performance \cite{ramirez2023explainable}.

\section{Results}

\label{sc:results}

\subsection{1:Best-Performing Architecture}
Our first analysis compares the performances of the eight NNs selected, both in the case of classification and regression tasks which are reported in \autoref{tab:my-table} across different sensor combinations. 
The combined results indicate that there is a statistically significant difference between the performance achieved with magnetoinertial (IMU) sensors and that with electromyography (EMG) sensors ($p=3.35*10^{-5}, p=2.20*10^{-3}, p=2.98*10^{-5}, p=8.32*10^{-4} $). Across all performance metrics—accuracy, precision, recall, and F1 score—the IMU sensor configuration consistently outperforms the EMG sensor configuration. For example, architectures such as CNN and LSTM show higher values when using IMU data, which likely reflects the robustness of the kinematic information (such as acceleration and angular velocity) captured by these sensors. In contrast, EMG signals, which measure muscle activation, tend to be more susceptible to variability due to factors like electrode placement, skin impedance, and signal cross-talk, leading to a noisier performance profile.
Moreover, the standard deviations observed with EMG data are generally higher, reinforcing the notion that its measurements are less reliable under the conditions tested. When comparing the IMU-only setting to the combined IMU+EMG setting, the differences in performance are negligible across all metrics. This lack of statistically significant improvement suggests that the inclusion of EMG data does not contribute meaningful additional information beyond what is already captured by the IMU sensors. The inertial data appears to encapsulate the essential dynamic features needed for accurate classification, thereby rendering the additional complexity of EMG integration unnecessary.
It is also possible that the nature of the movements being classified is such that the gross motion captured by the IMU is sufficiently discriminative, reducing the potential benefits of fusing EMG data. The redundancy in the information provided by the EMG signals may not offer any extra value in this specific scenario, especially when considering the potential for increased noise and processing overhead. In summary, these results support the conclusion that the IMU sensors are adequate for the intended application, providing robust performance without the need for additional EMG data.
The LSTM architecture consistently outperformed the other models across all evaluated performance metrics. In particular, for both IMU and IMU+EMG configurations, the LSTM achieved higher accuracy, precision, recall, and F1 scores compared to the second best performing architecture (blue in table). The computed p-value ($p=0.0015$, $p=0.0018$, $p=0.0014$, $p=0.0016$) confirms that this improvement is statistically significant and not merely a result of random variation. This robust performance can be attributed to the LSTM's inherent ability to capture long-term dependencies in sequential data. Its memory cells effectively filter noise and model temporal dynamics, which is especially important in handling the complex patterns found in sensor signals.

Moreover, the LSTM's architecture allows it to generalize better by retaining relevant contextual information over time, a feature that is crucial for accurately classifying time-series data. This capability gives it an edge over architectures such as CNNs, which are generally better at spatial feature extraction but may fall short in capturing temporal correlations. The lower variability in performance metrics for the LSTM further supports the idea that its superiority is intrinsic to its design rather than due to chance. Consequently, the statistically significant improvement, as evidenced by the p-value, highlights the robustness and effectiveness of the LSTM for this classification task.

\begin{table}[h!]
\centering
\caption{Classification Results: Each cell reports the average score followed by standard deviation. In black bold and blue bold the best and second-best results, respectively.}
\label{tab:my-table}
\resizebox{\textwidth}{!}{%
\begin{tabular}{|c|ccc|ccc|ccc|ccc|}
\hline
\multicolumn{1}{|c|}{} & \multicolumn{3}{c|}{\textbf{Accuracy}} & \multicolumn{3}{c|}{\textbf{Precision}} & \multicolumn{3}{c|}{\textbf{Recall}} & \multicolumn{3}{c|}{\textbf{F1 Score}} \\ \hline
\multicolumn{1}{|c|}{\textbf{Architecture}} & 
\textbf{IMU} & \textbf{EMG} & \textbf{IMU + EMG} & 
\textbf{IMU} & \textbf{EMG} & \textbf{IMU + EMG} & 
\textbf{IMU} & \textbf{EMG} & \textbf{IMU + EMG} & 
\textbf{IMU} & \textbf{EMG} & \textbf{IMU + EMG} \\ \hline
\multicolumn{1}{|c|}{CNN}      & $0.92\pm0.05$ & $0.73\pm0.10$ & $0.92\pm0.04$ & $0.92\pm0.03$ & $0.79\pm0.05$ & $0.92\pm0.06$ & $0.89\pm0.08$ & $0.73\pm0.04$ & $0.91\pm0.06$ & $0.89\pm0.05$ & $0.74\pm0.07$ & $0.91\pm0.06$ \\ \hline
\multicolumn{1}{|c|}{LSTM}     & $\mathbf{0.94\pm0.04}$ & $0.60\pm0.06$ & $\mathbf{0.94\pm0.03}$ & $\mathbf{0.95\pm0.04}$ & $0.39\pm0.07$ & $\mathbf{0.95\pm0.05}$ & $\mathbf{0.94\pm0.06}$ & $0.60\pm0.07$ & $\mathbf{0.94\pm0.08}$ & $\mathbf{0.94\pm0.05}$ & $0.46\pm0.07$ & $\mathbf{0.94\pm0.06}$ \\ \hline
\multicolumn{1}{|c|}{CNN-LSTM} & $0.90\pm0.05$ & $0.75\pm0.08$ & $0.93\pm0.04$ & $\textcolor{blue}{0.93\pm0.04}$ & $\mathbf{0.80\pm0.04}$ & $0.93\pm0.07$ & $\textcolor{blue}{0.92\pm0.05}$ & $\mathbf{0.75\pm0.03}$ & $0.90\pm0.07$ & $0.92\pm0.08$ & $\mathbf{0.76\pm0.03}$ & $0.91\pm0.07$ \\ \hline
\multicolumn{1}{|c|}{LSTM-CNN} & $0.91\pm0.05$ & $0.75\pm0.10$ & $0.91\pm0.04$ & $0.93\pm0.05$ & $0.78\pm0.04$ & $0.92\pm0.08$ & $0.91\pm0.03$ & $0.75\pm0.09$ & $0.90\pm0.06$ & $0.91\pm0.04$ & $0.76\pm0.06$ & $0.90\pm0.05$ \\ \hline
\multicolumn{1}{|c|}{GRU}      & $0.78\pm0.08$ & $0.61\pm0.03$ & $0.79\pm0.07$ & $0.78\pm0.07$ & $0.61\pm0.02$ & $0.78\pm0.06$ & $0.77\pm0.08$ & $0.60\pm0.03$ & $0.77\pm0.06$ & $0.79\pm0.07$ & $0.60\pm0.03$ & $0.79\pm0.10$ \\ \hline
\multicolumn{1}{|c|}{TST}      & $0.92\pm0.04$ & $0.74\pm0.03$ & $0.91\pm0.05$ & $0.89\pm0.08$ & $0.60\pm0.03$ & $0.89\pm0.06$ & $0.92\pm0.04$ & $0.74\pm0.02$ & $0.91\pm0.04$ & $\textcolor{blue}{0.92\pm0.05}$ & $0.74\pm0.03$ & $0.91\pm0.05$ \\ \hline
\multicolumn{1}{|c|}{XceptionTime} & $0.88\pm0.09$ & $0.73\pm0.03$ & $0.89\pm0.09$ & $0.88\pm0.09$ & $0.73\pm0.03$ & $0.88\pm0.01$ & $0.90\pm0.05$ & $0.74\pm0.03$ & $0.89\pm0.05$ & $0.88\pm0.09$ & $0.73\pm0.03$ & $0.88\pm0.09$ \\ \hline
\multicolumn{1}{|c|}{MAMBA}    & $\textcolor{blue}{0.92\pm0.04}$ & $\mathbf{0.76\pm0.05}$ & $0.91\pm0.06$ & $0.92\pm0.05$ & $0.75\pm0.05$ & $0.91\pm0.05$ & $0.89\pm0.05$ & $0.73\pm0.05$ & $0.91\pm0.05$ & $0.89\pm0.05$ & $0.75\pm0.05$ & $0.92\pm0.05$ \\ \hline
\multicolumn{1}{|c|}{Mean}     & $0.89\pm0.05$ & $0.71\pm0.06$ & $0.90\pm0.05$ & $0.90\pm0.05$ & $0.70\pm0.05$ & $0.90\pm0.05$ & $0.89\pm0.06$ & $0.71\pm0.06$ & $0.89\pm0.07$ & $0.89\pm0.06$ & $0.78\pm0.05$ & $0.89\pm0.07$ \\ \hline
\end{tabular}%
}
\end{table}

The average training time of the neural network architecture is about 21.25 second per epochs, with a range going from $11.42 \pm 0.15$ to $55.19 \pm 0.57$. On the other hand, all models demonstrated an inference time of \SI{1}{\milli\second} over a \SI{500}{\milli\second} lookback window.

In summary, our analysis demonstrates that IMU sensors consistently outperform EMG sensors, likely due to their robust capture of dynamic kinematic information. Furthermore, the negligible differences between the IMU-only and IMU+EMG configurations indicate that adding EMG data does not provide a significant advantage. Among the eight architectures evaluated, the LSTM emerged as the top performer.
For this reason, the regression analysis was conducted solely with IMU data, as the architectures showed no benefit when incorporating EMG signals, aligning with our aim to minimize the sensor setup.

Moreover, \autoref{tab:slope} The table demonstrates that the LSTM architecture achieves an MAE of 
for slope prediction, which is not only the lowest among all architectures but also statistically significantly different from the second-best performer (in blue in table). This significance confirmed by computed p-values ($p=0.015$) well below the standard threshold—indicates that the improvement is not due to chance. The LSTM's ability to capture long-term temporal dependencies and effectively filter noise in sequential data likely underpins this robust performance. Consequently, its lower error is a direct result of its architectural strengths rather than random variation.

\begin{table}[h!]
\centering
\caption{{Regression Analysis of Slope Inclination: MAE Performance.In black bold and blue bold the best and second-best results, respectively }}
\label{tab:slope}
\resizebox{0.4\textwidth}{!}{%
\begin{tabular}{|c|c|}
\hline
\textbf{Architecture} & \textbf{MAE [$^\circ$] (Slope)} \\ \hline
CNN          & $2.21\pm0.58$ \\ \hline
LSTM         & $\mathbf{1.93\pm0.53}$ \\ \hline
CNN-LSTM     & $2.10\pm0.68$ \\ \hline
LSTM-CNN     & $2.14\pm0.73$ \\ \hline
GRU          & $2.32\pm0.61$ \\ \hline
TST          & $2.21\pm0.60$ \\ \hline
XceptionTime & $\textcolor{blue}{2.08\pm0.54}$ \\ \hline
Mamba        & $2.03\pm0.46$ \\ \hline
Mean         & $2.12\pm0.59$ \\ \hline
\end{tabular}%
}
\end{table}
On average, the training process takes $24.00 \pm 0.33$ seconds per epoch for slope prediction ranging from $13.57 \pm 0.17$ to $64.34 \pm 0.23$. The inference time remains consistent across all models for both classification and regression tasks, averaging 1 ms.

Lastly, the table \autoref{tab:height} indicates that the CNN-LSTM architecture achieves the lowest MAE for stair height prediction, at $15.65\pm6.33mm$, outperforming the other architectures. In particular, its performance is statistically significantly better than that of the second-best architecture (LSTM, which has an MAE of 
$15.97 \pm 6.29mm$), as confirmed by the computed p-values  ($p=0.32$). This robust improvement is likely due to the CNN-LSTM's ability to combine the spatial feature extraction capabilities of CNNs with the temporal modeling strengths of LSTMs. By integrating these two approaches, the network is better equipped to capture the complex patterns and variations in stair height measurements, leading to a more accurate regression outcome that is not attributable to random chance.
\begin{table}[h!]
\centering
\caption{{Regression Analysis of Stair Height: MAE Performance. In black bold and blue bold the best and second-best results, respectively}}
\label{tab:height}
\resizebox{0.4\textwidth}{!}{%
\begin{tabular}{|c|c|}
\hline
\textbf{Architecture} & \textbf{MAE [mm] (Height)} \\ \hline
CNN          & $18.00\pm8.51$ \\ \hline
LSTM         & $\textcolor{blue}{15.97\pm6.29}$ \\ \hline
CNN-LSTM     & $\mathbf{15.65\pm6.33}$ \\ \hline
LSTM-CNN     & $16.05\pm7.40$ \\ \hline
GRU          & $21.10\pm9.11$ \\ \hline
TST          & $17.56\pm6.65$ \\ \hline
XceptionTime & $17.30\pm7.57$ \\ \hline
Mamba        & $16.02\pm6.89$ \\ \hline
Mean         & $17.20\pm7.34$ \\ \hline
\end{tabular}%
}
\end{table}
On average, the training process takes $20.14 \pm 0.33$ seconds per epoch for slope prediction ranging from $10.09 \pm 0.36$ to $58.45 \pm 0.36$. The inference time remains consistent across all models for both classification and regression tasks, averaging 1 ms.

\subsection{\textbf{2}: Most Informative Sensor}

In this section, we delve deeper into the analysis of the results, focusing on identifying the most informative sensors using the SHAP methodology. Building on our previous findings, we conduct this analysis exclusively with IMU data and the LSTM and CNN-LSTM model to enhance interpretability and insight.

\begin{figure}[h!]
    \centering
    \includegraphics[width=\textwidth]{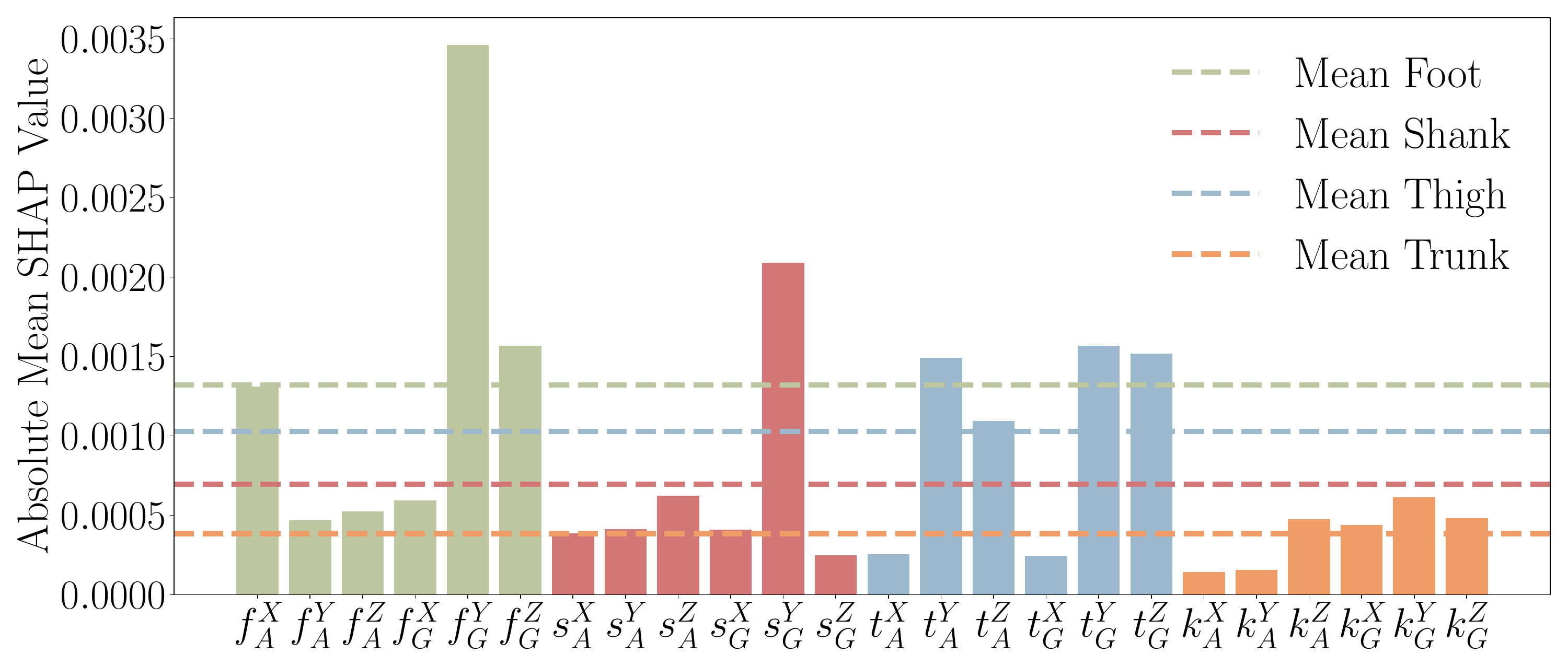}
    \caption{Feature importance for the IMU modality, derived by explaining the LSTM model used for the classification task with the SHAP methodology. 
    The labels x-axis are formatted as \( B_S^A \), where \( B \) denotes the body part (foot: \( f \), shank: \( s \), thigh: \( t \), trunk: \( k \)), \( S \) indicates the sensor type (accelerometer: \( A \), gyroscope: \( G \)), and \( A \) specifies the axis (x-axis: \( X \), y-axis: \( Y \), z-axis: \( Z \)). 
    The dotted horizontal lines represent the average sensor weights.
}
    \label{fig:imu_shap}
\end{figure}

~\autoref{fig:imu_shap} shows how much each sensor influences the ground type prediction where f, s, t, and k represent the foot, shank, thigh, and trunk, respectively. Additionally, the superscript denotes the axis (x, y, or z), while the subscript indicates the sensor type: G for the gyroscope and A for the accelerometer. The SHAP analysis indicates that foot sensors have the greatest influence on ground type prediction, followed by those on the thigh, shank, and trunk. This ranking aligns closely with the biomechanics of human locomotion and how different body segments interact with the ground.

It is evident that the foot-mounted IMU provides the most informative kinematic data for terrain classification, as the foot undergoes the most significant motion variations during locomotion. Being the primary point of contact with the ground, it is logical to deduce that the foot experiences changes in acceleration and angular velocity that directly reflect terrain properties \cite{shi2024simple}. Step transitions, foot placement, and ankle dynamics induce distinctive kinematic signatures, which are more pronounced than those captured by sensors positioned elsewhere on the body. Consequently, a Foot-mounted IMU is capable of detecting terrain-induced variations more effectively, making them the optimal choice for extracting kinematic features relevant to surface characterization.

IMU sensors positioned on the shank and thigh are capable of capturing key kinematic adaptations to terrain changes, though their contribution is less pronounced than foot-mounted sensors. Variations in surface incline or height have been shown to influence knee flexion, stride length, and hip motion, leading to distinct acceleration and angular velocity patterns. These segments have been demonstrated to play a role in adjusting limb trajectories and modulating propulsion, particularly during slope ascent, descent, or stair negotiation. 

In contrast, trunk-mounted sensors provide the least informative kinematic data for terrain classification. While the trunk stabilizes movement and reacts to lower-limb dynamics, its motion is less directly affected by ground properties. As a result, kinematic variations measured at the trunk are less pronounced, leading to lower relevance for identifying terrain characteristics compared to sensors positioned on the foot and lower limbs.

The observed sensor hierarchy is consistent with how humans adjust their movement to different terrains. Foot sensors capture direct interactions with the ground, thus providing the most informative data, while the shank and thigh assist in adjusting movement. The trunk plays a more supportive role in stabilization than in direct terrain response, which leads to its lower contribution to ground prediction.

\begin{figure}[h!]
    \centering
    \includegraphics[width=\textwidth]{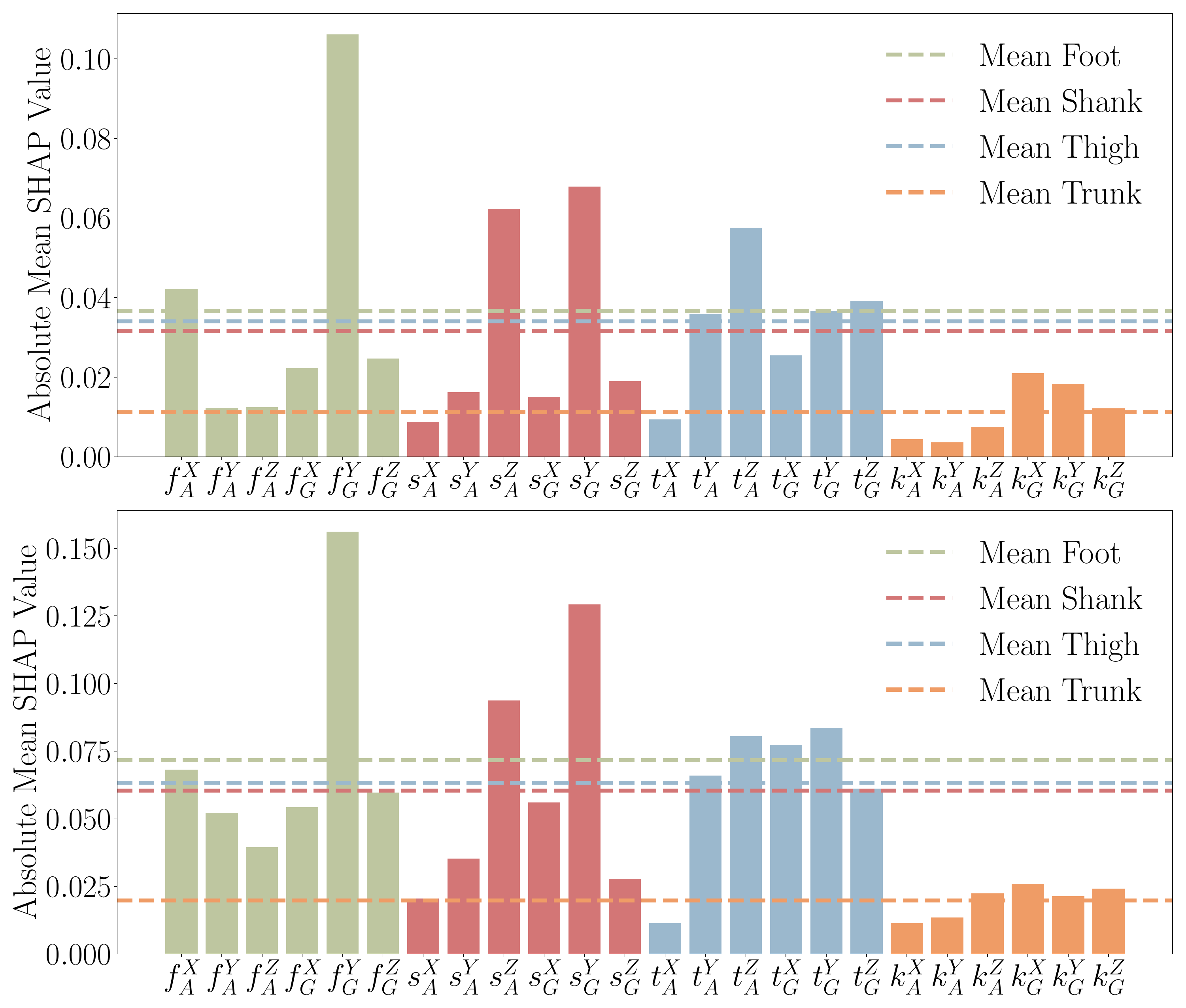}
    \caption{Feature importance for the IMU modality derived using the SHAP methodology, with the upper plot showing the LSTM model for slope prediction and the lower plot showing the CNN-LSTM model for stair height prediction, respectively.
    The features on x-axis are formatted as \( B_S^A \), where \( B \) denotes the body part (foot: \( f \), shank: \( s \), thigh: \( t \), trunk: \( k \)), \( S \) indicates the sensor type (accelerometer: \( A \), gyroscope: \( G \)), and \( A \) specifies the axis (x-axis: \( X \), y-axis: \( Y \), z-axis: \( Z \)). 
    The horizontal lines represent the average sensor weights.
    }
    \label{fig:shap-ramp+scale}
\end{figure}

Focusing on the two regression tasks,~\autoref{fig:shap-ramp+scale} adopts the same notation and shows the results of the SHAP analysis both for slope prediction using LSTM (upper panel) and for stair height prediction using the CNN-LSTM model (lower panel). Both plots confirm that foot sensors are the most influential of these two regressions, with the shank and thigh sensors following closely. 

As expected, the trunk sensor consistently contributed the least. These insights underscore the critical role of foot sensors in capturing essential gyroscope and acceleration data during gait~\cite{prasanth2021wearable}. Also in this case, analyzing each sensor's features, we see that the most influential is consistently the gyroscope on the y-axis of the foot. In second place is the thigh, where all sensors contribute equally except for the accelerometer on the x-axis. For what concerns the Shank the two most influential features are the acceleration on the z-axis and the gyroscope on the y-axis. The results indicate that the gyroscope on the y-axis of the foot is the most influential feature for predicting stair height and slope inclination, which is quite expected. This makes sense because the y-axis of the foot gyroscope captures rotational movement in the sagittal plane, which directly correlates with stair-climbing movements and slope changes.
For the thigh, the fact that all sensors contribute equally—except for the accelerometer on the x-axis—is also reasonable. The thigh experiences both rotational and linear movements during stair ascent and descent, making all sensor modalities relevant. The lower influence of the x-axis accelerometer suggests that lateral movements of the thigh are less critical in determining stair height or slope.
Regarding the shank, the dominance of the z-axis acceleration and the y-axis gyroscope aligns with expectations. The z-axis acceleration captures vertical displacement and impact forces, which are crucial for stair-related tasks. Meanwhile, the y-axis gyroscope reflects rotational motion in the sagittal plane, which is heavily involved in the adaptation to different stair heights and inclinations.
Overall, these results align well with the kinematics of stair negotiation, where foot and shank dynamics play a critical role in adapting to slope and step height, while thigh movement remains more evenly distributed across sensors. The findings reinforce the importance of gyroscopic data, particularly along the y-axis, in capturing stair-climbing mechanics.

The previous analysis led us to reassess our sensor setup to simplify it without compromising performance. 
We evaluated four sensor combinations, starting with one IMU sensor and progressively adding sensors placed on different body districts based on their SHAP-derived importance.
This iterative retraining process allowed us to assess the impact of each sensor combination on classification and regression performance.

\begin{figure}[h]
    \centering
    \includegraphics[width=\textwidth]{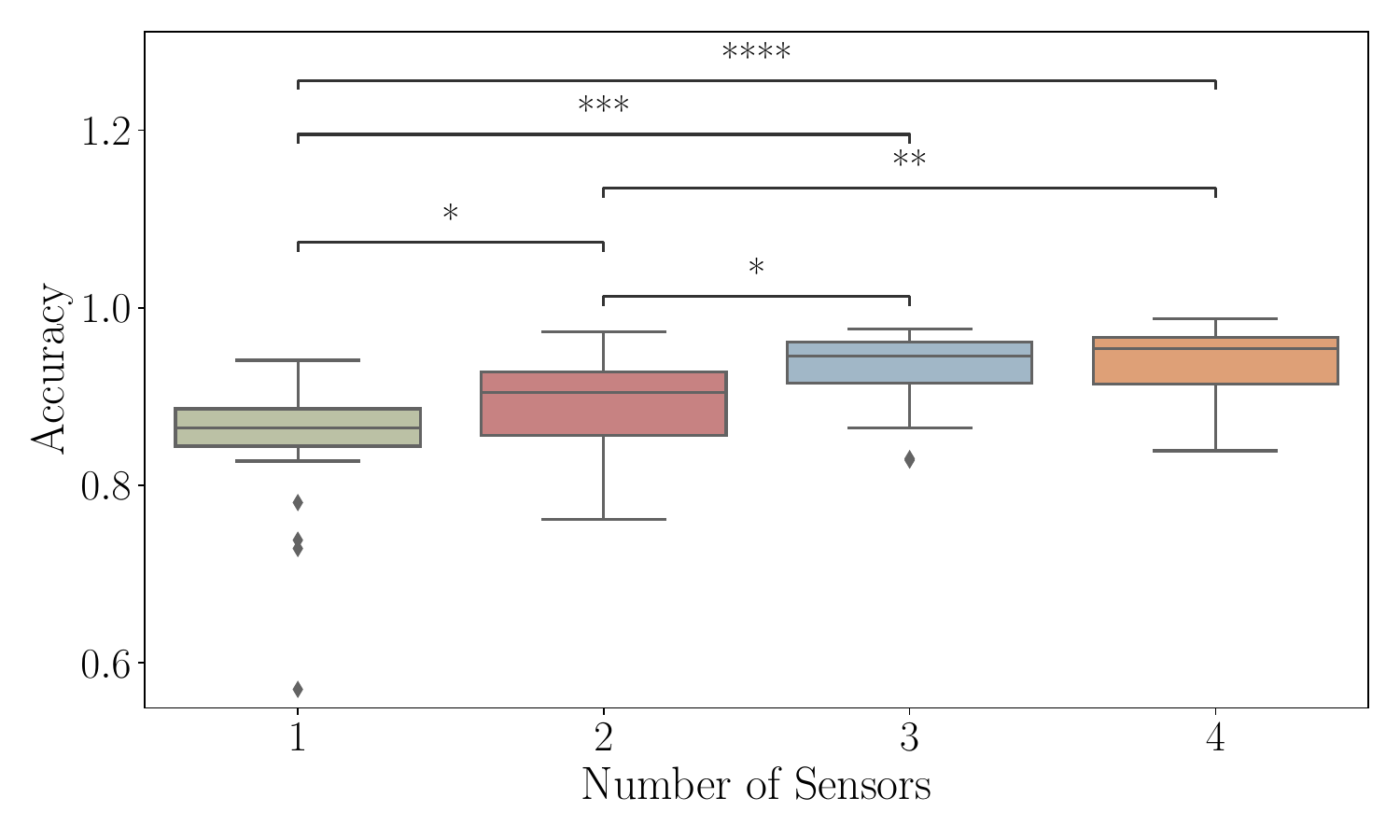}
    \caption{Accuracy of the LSTM model in the classification task as a function of the number of IMU sensors iteratively added based on their informativeness ranked by the SHAP methodology across the four body sectors.
    Statistical significance is denoted by asterisks, with * indicating \(p \leq 0.05\), ** indicating \(p \leq 0.01\), *** indicating \(p \leq 0.001\), and **** indicating \(p \leq 0.0001\).}
    \label{fig:sensor_configuration}
\end{figure}

\autoref{fig:sensor_configuration} illustrates the classification accuracy of the LSTM model as a function of the IMU sensors considered. The results reveal a statistically significant difference between setups with four sensors and those with one or two sensors. However, there was no significant difference between using three sensors versus four, suggesting that the trunk sensor does not substantially contribute to system functionality. 
Consequently, reducing the sensor setup to three sensors (foot, thigh, and shank) does not result in a significant performance drop, offering a more practical and cost-effective configuration.

\begin{figure}[http]
    \centering
    \includegraphics[width=\textwidth]{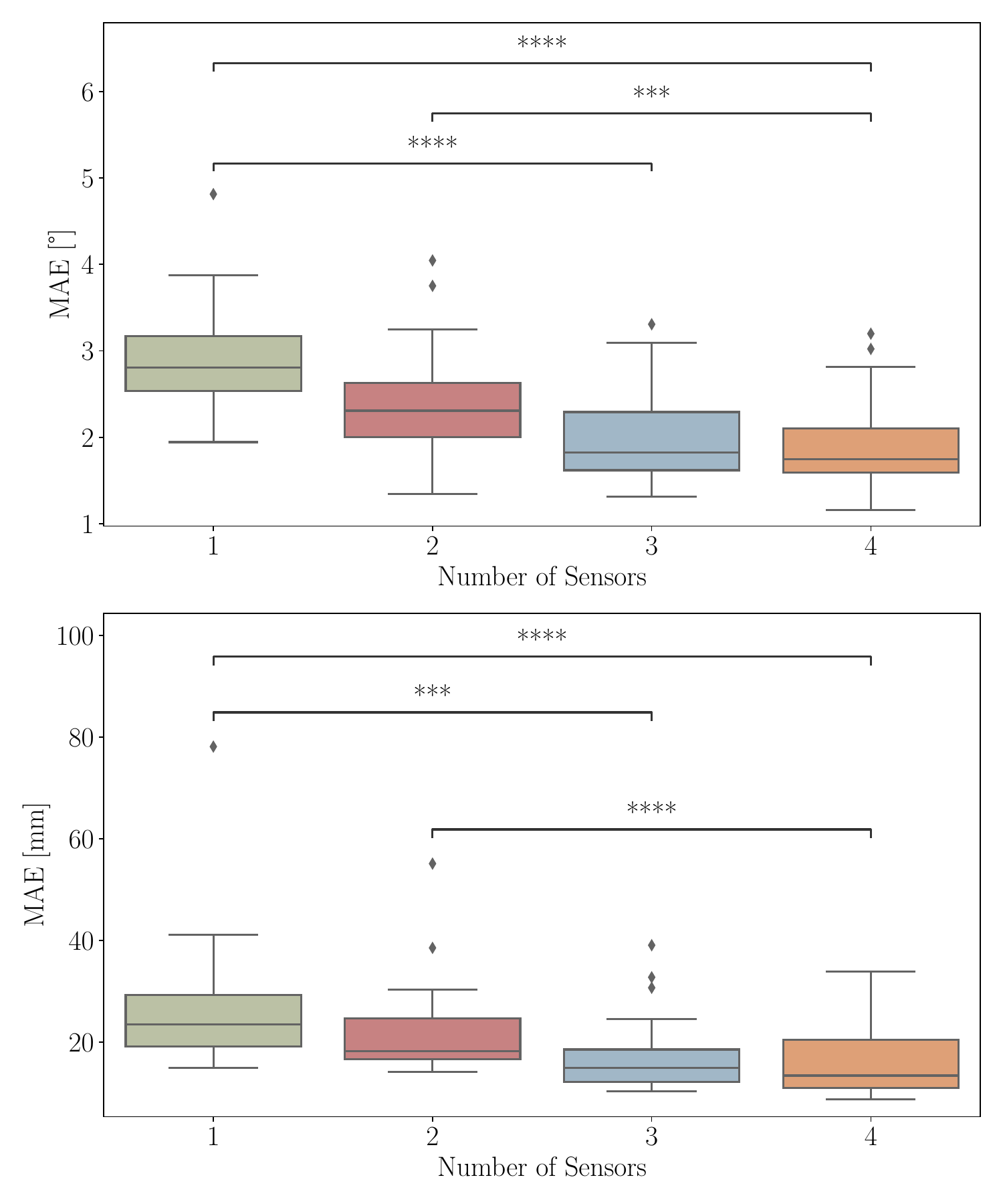}
    \caption{The MAE metric as a function of the number of IMU sensors iteratively added based on their informativeness ranked by the SHAP methodology across the four body sectors, with the upper plot showing slope prediction case and the lower plot showing stair height prediction case.  Statistical significance is denoted by asterisks, with *** indicating \(p \leq 0.001\), and **** indicating \(p \leq 0.0001\).
    }
    \label{fig:mae_sensors}
\end{figure}

Similarly, \autoref{fig:mae_sensors} presents the MAE as a function of the IMU sensors, with sensors added iteratively based on their SHAP-derived importance for slope prediction (upper plot) and stair height prediction (lower plot). 
The findings indicate that excluding the trunk sensor did not significantly affect performance in either regression task, supporting the conclusion that we can exclude it from the setup.

\section{Discussion}
This study provides a comprehensive comparative analysis of deep learning models applied to human locomotion classification and terrain parameter estimation, particularly for lower-limb robotic exoskeletons. The results demonstrate that different architectures excel in locomotion modelling with and accuracy higher than 0.90, with LSTM emerging as the most effective model overall. However, a hybrid CNN-LSTM approach proves superior for stair height regression, highlighting the nuanced relationship between spatial and temporal dependencies in movement analysis.

LSTM’s strength lies in its ability to capture long-term dependencies in sequential data, a fundamental requirement for understanding locomotion patterns. Unlike CNN, which is highly effective at learning spatial representations but lacks an inherent mechanism for handling temporal relationships, LSTM processes sequences by maintaining a memory of past information through its gated architecture. These gates regulate the flow of information, allowing the network to retain relevant features while discarding noise, which is particularly useful in gait analysis where movements evolve dynamically over time. For ground-level classification, LSTM demonstrates the highest performance due to its ability to recognize movement transitions and maintain continuity in gait sequences, making it well-suited for distinguishing between different locomotion modes.

In contrast, stair height regression benefits more from the CNN-LSTM hybrid model, which integrates the advantages of both convolutional and recurrent architectures. Stair climbing involves distinct movement phases, including lifting, pushing off, and landing, where CNN effectively extracts localized features related to these abrupt transitions. By combining CNN with LSTM, the model retains the ability to track the step-by-step progression while also refining the height estimation through convolutional feature extraction. This hybrid approach ensures that both spatial variations and long-term dependencies are adequately modeled, leading to more accurate predictions of stair height compared to an LSTM-only approach.

For slope regression, however, LSTM again outperforms other models, suggesting that gradual changes in movement mechanics are best captured by recurrent architectures. Unlike stair climbing, where discrete transitions between steps occur, slope walking involves a continuous evolution of movement patterns. The recurrent design of LSTM allows it to process these gradual shifts effectively, preserving the relationship between past and present gait states without the constraints imposed by convolutional layers. While transformers, such as the Time Series Transformer (TST) and MAMBA, offer alternative paradigms for sequential modeling through self-attention mechanisms, their advantage is more pronounced in datasets with extremely long dependencies. In gait analysis, where temporal relationships are structured and periodic, LSTM remains a more efficient and reliable solution.

The study also examined the performance of MAMBA, XceptionTime, and GRU, which were found to be less effective in comparison to LSTM and CNN-LSTM. MAMBA, a state-space model designed to handle sequential data without recurrence, struggles with locomotion analysis due to its reliance on a different computational framework that may not be as finely tuned to the biomechanical constraints of human movement. While MAMBA excels in handling general sequential patterns, it lacks the ability to preserve structured periodic dependencies crucial for gait modeling. XceptionTime, a convolution-based model adapted for time-series tasks, faces inherent limitations due to its primary reliance on spatial feature extraction rather than the sequential nature of locomotion data. While it can learn feature hierarchies effectively, its inability to capture temporal transitions over time reduces its effectiveness in gait analysis. Similarly, GRU, despite being a recurrent model like LSTM, operates with a simpler gating mechanism that makes it computationally efficient but potentially less expressive in handling complex sequential patterns. GRU is effective at modeling short-term dependencies but may not retain sufficient context for more nuanced motion patterns that evolve over longer time spans, which explains why LSTM remains preferable in this scenario.

Beyond model selection, the study also emphasizes the critical role of sensor configuration. Although it was initially expected that a multimodal approach incorporating both IMU and EMG data would yield superior performance due to the additional physiological information from muscle activity, the findings indicate otherwise. IMU data alone proves sufficient and even more effective, suggesting that the richness of acceleration and gyroscope signals in capturing movement dynamics outweighs the benefits of integrating EMG. The additional complexity introduced by EMG signals does not lead to a significant performance gain, reinforcing the idea that simpler sensor setups can be both practical and highly effective. Moreover, an analysis of feature importance using SHAP confirms that a minimal yet strategic sensor placement—covering the foot, shank, and thigh—achieves an optimal balance between accuracy and usability.

The findings of this study carry significant implications for wearable robotic exoskeletons working in unstructured environments. The ability to classify ground conditions and estimate locomotion parameters with high accuracy and minimal latency would directly influence the adaptability, safety, and usability of such systems. In real-world applications, exoskeletons should continuously interpret the environment and anticipate transitions to ensure seamless support, thereby preventing the user's destabilization or excessive cognitive burden.
The ability to reliably distinguish between different locomotion contexts (e.g. level ground, ramps, stairs) would enable exoskeletons to preemptively adjust assistance strategies, optimizing the trajectories, the interaction control, and step timing. This is particularly crucial in rehabilitation, where precise adaptation to terrain variations can enhance training efficacy by encouraging more natural and effortful gait patterns. In the context of assistive applications for individuals with mobility impairments, real-time ground adaptation reduces the need for manual intervention, thereby fostering greater autonomy and reducing user fatigue. 

Furthermore, the accuracy of parameter estimation (e.g., stair height, ramp inclination) impacts the granularity of control adjustments. A mismatch between perceived and actual locomotion demands could lead to either overcompensation, resulting in unnatural and inefficient movements, or undercompensation, which may compromise stability. The accuracy obtained by the proposed approach suggests that fine-grained terrain adaptation is feasible, paving the way for exoskeletons that can seamlessly modulate assistance not only across different terrains but also in response to subtle environmental variations.

The findings of this study serve to reinforce the critical role of real-time perception in the domain of wearable robotics. The reliable classification of terrain and the estimation of locomotion parameters are not merely technical benchmarks; they are fundamental enablers of more intuitive, responsive, and independent mobility assistance, thereby bridging the gap between robotic exoskeletons and natural human movement.

\section{Conclusions}
\label{sc:concl}
In this study, we have presented a neural network-based system for real-time ground condition analysis, finding that the integration of three specialized models may enhance adaptive robotic exoskeleton control. Specifically, an LSTM classifier identifies terrain type across five categories, a second LSTM estimates ramp slope, and a hybrid CNN-LSTM predicts stair height. A comparative analysis of literature models trained on the public CAMARGO dataset, together with SHAP explanations, informed the final architecture, which relies on three IMU sensors positioned on the foot, shank, and thigh to minimize sensor burden while maintaining high accuracy.
Furthermore, the use of SHAP for sensor relevance assessment revealed that the trunk sensor contributed minimally to both classification and regression tasks, enabling its removal without degrading performance. This streamlined setup maintained high accuracy in terrain classification (0.94 ± 0.04), as well as precise estimations of ramp slope (1.95 ± 0.58°) and stair height (15.65 ± 7.40 mm), corroborating findings in the literature. Notably, inference times of approximately 1 ms make the system suitable for real-time applications, such as lower-limb exoskeleton control.
Contrary to the widespread assumption that multimodal sensor inputs improve performance, our experiments demonstrated that IMU data alone outperformed an IMU+EMG combination, reinforcing the efficiency of an IMU-only approach. These findings support a cost-effective and lightweight sensor configuration while maintaining robust classification and regression performance.
\\Future work will focus on validating the proposed system in real-world scenarios with healthy participants and developing control strategies for lower-limb exoskeletons that integrate the predicted locomotion parameters. This research contributes to the advancement of intelligent exoskeletons, offering accurate, low-latency terrain awareness for adaptive control in practical applications.

\section{Aknowledgment}
Coser Omar is a Ph.D. student enrolled in the National Ph.D. in Artificial Intelligence, XXXVIII cycle, course on Health and life sciences, organized
by Università Campus Bio-Medico di Roma.
Resources are partially provided by the National Academic Infrastructure for Supercomputing in Sweden (NAISS) and the Swedish National Inffrastructure for Computing (SNIC) at Alvis @ C3SE, partially funded by the swedish Research Council through grant agreement no 2022-06725 and no. 2018-05973


\end{document}